\PassOptionsToPackage{dvipsnames}{xcolor}
\documentclass{lab}
\usepackage[utf8]{inputenc}
\usepackage[T1]{fontenc}
\usepackage{lmodern}
\usepackage{amsmath,amssymb}
\usepackage{charter}
\usepackage{xurl}
\usepackage[shortlabels,inline]{enumitem}

\usepackage{xspace}
\newcommand{\ours}{Entangle\xspace}

\usepackage{booktabs}
\usepackage{colortbl}
\usepackage{tabularx}
\usepackage{multirow}
\definecolor{tableRoPE}{HTML}{EFF8F0}
\definecolor{tableContext}{HTML}{FCFBEA}
\definecolor{tableOurs}{HTML}{EDF1FC}
\usepackage{placeins}

\usepackage{amsmath,amsfonts,bm}

\def\eqref#1{equation~\ref{#1}}

\def\1{\bm{1}}

\DeclareMathAlphabet{\mathsfit}{\encodingdefault}{\sfdefault}{m}{sl}
\SetMathAlphabet{\mathsfit}{bold}{\encodingdefault}{\sfdefault}{bx}{n}

\usepackage{longtable}
\usepackage{needspace,wrapfig}

\setlist[itemize]{noitemsep,leftmargin=*,topsep=0em}
\setlist[enumerate]{noitemsep,leftmargin=*,topsep=0em}

\title{\ours: Scaling Human-Video In-Context World-Action Models across Unseen Tasks and Contexts}
\renewcommand{\authorlist}{%
  {\authorfont
    \authorformat[*]{Chendong Xiang},
    \authorformat[*]{Chengwen Zhang},
    \authorformat{Ruowen Zhao},
    \authorformat{Zeyuan Wang},
    \authorformat{Shuhe Huang}\par
    \authorformat{Rongxu Cui},
    \authorformat{Min Zhao},
    \authorformat{Fan Bao},
    \authorformat{Hengkai Tan},
    \authorformat{Jun Zhu}\par
  }%
}
\affiliation{Tsinghua University, GensPI Technology, Motubrain}
\newcommand{\equalcontributionfootnote}{%
  \begingroup
    \renewcommand{\thefootnote}{\fnsymbol{footnote}}%
    \footnotetext[1]{These authors contributed equally and share first authorship.}%
  \endgroup
}

\hypersetup{
  pdftitle={Entangle: Scaling Human-Video In-Context World-Action Models across Unseen Tasks and Contexts},
  pdfauthor={Chendong Xiang, Chengwen Zhang, Ruowen Zhao, Zeyuan Wang, Shuhe Huang,
    Rongxu Cui, Min Zhao, Fan Bao, Hengkai Tan, Jun Zhu},
}

\begin{document}
\maketitle
\equalcontributionfootnote

\begin{tcolorbox}[
  enhanced,
  breakable,
  colback=labblue!10,
  colframe=labblue!35,
  boxrule=0.5pt,
  arc=6pt,
  left=10pt,
  right=10pt,
  top=2pt,
  bottom=2pt
]
\abstract{In-context learning (ICL) enables foundation models to perform unseen tasks from context without parameter updates.
In robotic manipulation, a human video provides the context that specifies how a task should be performed.
However, scaling human-video ICL on world-action models (WAMs) faces two challenges: task-level human--robot pairs are not temporally aligned with robot actions, and standard action losses allow policies to rely on initial-scene or task-identity shortcuts rather than tracking the context.
We present \textbf{\ours}, an in-context world-action model that maps a single human video and current robot observations to future visual states and executable actions.
To scale temporally aligned training data, we convert recorded robot videos into matched human contexts via video-to-video generation, supplemented by real paired recordings.
For model training, we combine variable-length video windows with an intra-task contrastive objective that distinguishes the paired action trajectory from other executions of the same task, preventing task-identity shortcuts.
Without task-specific fine-tuning, \ours improves steadily as aligned training data scales, reaching an average score of 83.5/100 across six unseen tasks on an Astribot S1 robot, and achieves a 69.5\% average success rate on seven unseen RoboTwin tasks, outperforming Zero-WAM by 22.6 percentage points.
Under Context-OOD, although trained only on scene-aligned pairs, \ours transfers context-specified behaviors across rearranged layouts and substituted tools.

\par\smallskip
\noindent\textbf{Project page.} \url{https://entangle-robotics.github.io/}
}
\end{tcolorbox}

\begin{minipage}{\textwidth}
  \centering
  \includegraphics[width=\textwidth]{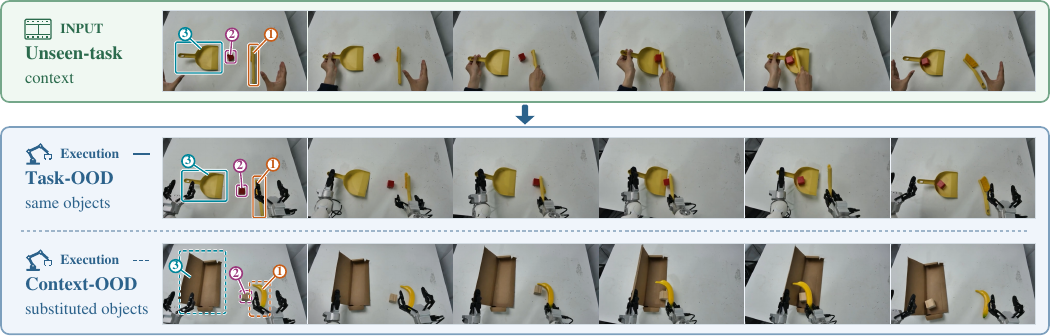}
  \captionsetup{hypcap=false}
  \captionof{figure}{\textbf{Executing an unseen task.}
  The input context (top) shows a human using a brush (1) to sweep a
  plastic block (2) into a dustpan (3), a task absent from training.
  \textbf{Task-OOD} (middle): conditioned on this context, the robot executes
  the unseen task with the same objects.
  \textbf{Context-OOD} (bottom): the robot executes the same interaction with
  objects that differ from those in the input context, using a banana (1) to sweep
  a wooden block (2) into a cardboard box (3);
  dashed outlines mark substituted objects.
  }
  \label{fig:teaser}
\end{minipage}
\par

\section{Introduction}
\label{sec:intro}

In-context learning (ICL) enables foundation models to perform unseen tasks from context without updating model parameters~\cite{brown2020language}.
In robotic manipulation, language instructions often underspecify multi-step order, spatial constraints, and fine-grained motion, whereas a single human demonstration video directly shows how a task should be performed~\cite{jain2024vid2robot,fu2025icrt,vosylius2025instant,sridhar2025ricl,zhang2026understanding}.
World-action models (WAMs)~\cite{ye2026world,zhang2026native,zhou2026zerowam,motubrain2026,vidarc2025}, which jointly model future visual dynamics and robot control within a shared generative backbone, offer a natural foundation for such video-conditioned policies: given a single human demonstration video as context, a WAM can predict future visual states and executable actions for unseen manipulation tasks without parameter updates.

However, scaling human-video ICL on WAMs faces two core challenges.
First, human and robot videos are typically paired only at the task level.
Whether collected independently or synthesized from an edited initial frame via text-to-video generation~\cite{zhou2026zerowam,bao2024vidu}, task-level pairs share a high-level goal but diverge in motion paths, contact timing, and execution speed.
Without temporal alignment between the human video and ground-truth robot actions, the model cannot localize which segment of the demonstration corresponds to a target action chunk, reducing the video prompt to a coarse task label.
Second, even when paired trajectories are aligned, standard action-prediction losses supervise each trajectory in isolation and leave room for shortcuts: when a task admits multiple valid executions (e.g., different manipulation orders or obstacle-avoidance paths), a policy can lower its loss by inferring the task identity or relying on the initial scene layout rather than tracking the specific progression shown in the demonstration.

To address these challenges, we present \textbf{\ours}, an in-context world-action model that maps a single human demonstration video and current robot observations to future visual states and executable actions.
First, to scale temporally aligned training data, we convert recorded robot videos into matched human demonstrations via reference-conditioned video-to-video (V2V) generation, supplemented by real paired recordings (Fig.~\ref{fig:paired_dataset_overview}).
Unlike first-frame editing followed by text-driven generation, V2V translation preserves the camera viewpoint, object interactions, and timing of the source robot trajectory, so each generated human video remains temporally aligned with the recorded ground-truth actions.
Second, we encode the human demonstration and current robot observations in a shared video-action backbone and concatenate their tokens along the temporal axis (Fig.~\ref{fig:model_architecture}).
For model training, we combine variable-length video windows with an intra-task InfoNCE contrastive objective~\cite{oord2018representation} inspired by negative-aware fine-tuning~\cite{zheng2026diffusionnft}: by pulling predicted actions toward the video-paired trajectory while repelling other trajectories of the \emph{same} task, it prevents task-identity shortcuts and forces the policy to follow the demonstrated motion.

We evaluate \ours on an Astribot S1 robot and the RoboTwin benchmark~\cite{mu2025robotwin,chen2025robotwin2}.
Without task-specific fine-tuning, \ours improves steadily as aligned training data scales, reaching an average score of 83.5/100 across six unseen tasks on Astribot S1, and achieves a 69.5\% average success rate on seven unseen RoboTwin tasks, outperforming Zero-WAM~\cite{zhou2026zerowam} by 22.6 percentage points.
More surprisingly, although our training pairs are strictly scene-aligned, \ours goes beyond unseen tasks with the same objects (Fig.~\ref{fig:teaser}, middle): it transfers the demonstrated behavior across rearranged layouts and substituted tools, such as using a banana to sweep a wooden block into a cardboard box from a brush-and-dustpan demonstration (Fig.~\ref{fig:teaser}, top and bottom).

Our main contributions are as follows:
\begin{itemize}
    \item \textbf{Scalable temporally aligned data:} we turn recorded robot trajectories into temporally matched human demonstrations with reference-conditioned video-to-video generation, so alignment scales with the robot-video corpus without human reenactment.
    \item \textbf{In-context WAM training:} we concatenate demonstration and robot-video tokens along time, train on variable-length windows, and add an intra-task \mbox{InfoNCE} loss that removes the task-identity shortcut.
    \item \textbf{Data scaling and cross-scene transfer:} unseen-task scores grow monotonically with aligned data; scene-aligned training transfers to new layouts and tools.
\end{itemize}

\section{Related Work}
\label{sec:related_work}

\subsection{In-Context Learning for Robotic Manipulation}
In-context learning enables a frozen robot policy to infer a task from deployment demonstrations, such as sensorimotor trajectories in ICRT~\cite{fu2025icrt}, demonstration graphs in Instant Policy~\cite{vosylius2025instant}, or retrieved segments in RICL~\cite{sridhar2025ricl}, extending one-shot policies based on task embeddings~\cite{duan2017oneshot,james2018tecnets}, frame attention~\cite{dasari2021transformers,mandi2022mosaic}, or skill priors~\cite{yu2019composition,xu2023xskill,zhu2026humanprompt}.
To leverage human videos, prior works learn visual features~\cite{pmlr-v205-nair23a}, latent actions~\cite{ye2025lapa}, or play plans~\cite{pmlr-v229-wang23a}, co-train on egocentric data in EgoMimic~\cite{kareer2025egomimic}, retrieve play segments in RHyME~\cite{kedia2025one} and MimicDroid~\cite{shah2026mimicdroid}, or condition deployment on a single human video via video retrieval in ViP~\cite{chanesane2023vip}, language grounding in BC-Z~\cite{pmlr-v164-jang22a}, and cross-attention in Vid2Robot~\cite{jain2024vid2robot}.
However, reducing a demonstration to a task identifier limits policies to recalling training behaviors~\cite{pan2023disentangling}, whereas generalizing to unseen procedures and novel object uses requires tracking \emph{how} the interaction unfolds, as explored through subgoal or trace prediction~\cite{goswami2026osvi,qian2026synthicl,son2026seetraceact}, temporal alignment~\cite{sermanet2018tcn,dwibedi2019tcc,zakka2022xirl,chen2026host}, intra-task keyframe separation in JEDI~\cite{liang2026jedi}, layout shifts in SimpleICL~\cite{li2026simpleicl}, and functional substitutions in The Imitator Game~\cite{zhou2026imitatorgame}.
\ours uses a single human video to specify the procedure within a world-action model, generalizing across unseen procedures (Task-OOD) and altered scenes or tools (Context-OOD).

\subsection{World-Action Models}
World-action models jointly predict future observations and actions~\cite{wu2024gr1,li2025unified,zhu2025unified,bi2026motus,motubrain2026}, extending visual foresight~\cite{finn2017foresight}, action recovery from visual plans~\cite{du2023learning,ko2024avdc,liang2024dreamitate}, predictive representations for control~\cite{tian2025seer,hu2025vpp,zheng2025flare}, and video generators adapted for policy learning~\cite{shen2026videovla,ye2026world,li2026causal,vidar2025,vidarc2025} or rollout synthesis~\cite{jang2025dreamgen}.
Recent WAMs further incorporate context and demonstrations: ICWM~\cite{wang2026context} infers system properties from self-interaction, WATCH~\cite{watch2026} pretrains on task-paired human videos, LingBot-VA 2.0~\cite{zhang2026native} and WorldScape Policy 2.0~\cite{su2026worldscape} accept demonstration prompts, and Zero-WAM~\cite{zhou2026zerowam} generates task-matched human clips while predicting temporally spaced future chunks to encourage prompt usage.
However, conditioning on a demonstration alone does not ensure that predicted actions follow the demonstrated execution over task-level cues; \ours addresses this by combining temporally aligned training with intra-task action contrast~\cite{oord2018representation} to distinguish different executions of the same task.

\subsection{Test-Time Training}
Meta-imitation adapts policy parameters from a demonstration~\cite{finn2017oneshot,yu2018one}, while test-time training adapts models or fast weights to test inputs~\cite{sun2020test,hansen2021policy,sun2025learning}.
WAM-TTT~\cite{feng2026wam} updates fast weights from human videos before rollout, and RoboTTT~\cite{jiang2026robottt} updates them as the robot's visuomotor history grows.
\ours keeps its weights fixed at deployment, conditioning each prediction directly on the human demonstration and current robot observation.

\subsection{Motion Retargeting and Paired Data}
Motion retargeting transfers object-relative motions~\cite{heppert2024ditto,zhu2024orion} or human hand trajectories~\cite{li2024okami,park2025demodiffusion,qin2022dexmv} to a robot; BEHAVE~\cite{bhatnagar2022behave} records human--object contact, and CORE4D~\cite{liu2025core4d} retargets two-person rearrangement motions to new object shapes.
Robot training data can also come from paired human--robot recordings~\cite{sharma2018mime,fang2024rh20t,human2robot2025} or synthesizing robot demonstrations from human footage~\cite{lepert2025phantom,egoengine2026,lee2026robotryon,chen2025vivla}, whereas Gen2Act~\cite{pmlr-v305-bharadhwaj25a} generates a human video from an initial image and task description.
\ours uses the full robot recording as a motion reference, verifies the generated human video for temporal and interaction agreement with the source, and combines these clips with synchronized human reenactments so each sampled demonstration window pairs directly with the recorded robot actions.

\section{Ent-Dataset Construction}
\label{sec:paired_dataset}

\subsection{Dataset Overview and Strict Pairing}
\label{sec:strict_pairing}

\begin{figure}[!htbp]
  \centering
  \includegraphics[width=\textwidth]{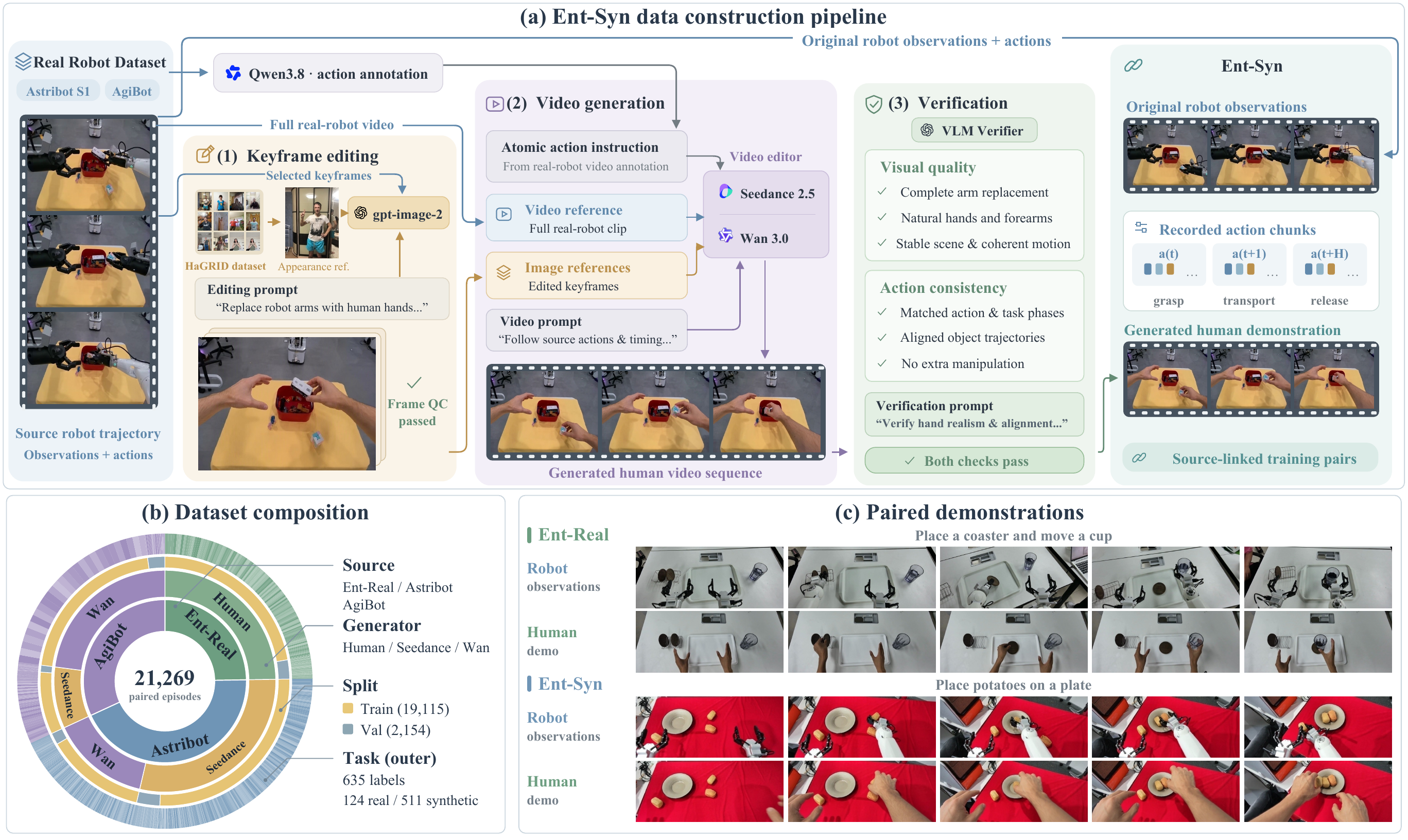}
  \caption{\textbf{Overview of the Ent-Dataset.}
  (a) \textbf{Ent-Syn pipeline:} source robot trajectories undergo (1) keyframe hand editing, (2) motion-guided video generation, and (3) VLM verification for visual quality and action consistency, before pairing the generated human video with the original robot observations and actions.
  (b) \textbf{Dataset composition:} concentric rings (from inner to outer) break down all 21,269 paired episodes by data source (Ent-Real, Astribot, AgiBot), human-video source (Human, Seedance, Wan), train/validation split (19,115 / 2,154), and task category (635 tasks).
  (c) \textbf{Paired demonstrations:} examples from the Ent-Real and Ent-Syn subsets across five matched timestamps, illustrating temporally aligned robot and human executions.}
  \label{fig:paired_dataset_overview}
\end{figure}

Each training example in the Ent-Dataset pairs a human demonstration video
with a recorded robot trajectory containing camera observations,
proprioceptive states, and actions (Fig.~\ref{fig:paired_dataset_overview}(c)).
As summarized in Fig.~\ref{fig:paired_dataset_overview}(b), the dataset
contains 21.3k pairs split across disjoint robot episodes for training and
validation, comprising 5.3k real pairs from Astribot S1 (\textbf{Ent-Real})
and 16.0k synthetic pairs from Astribot and AgiBot (\textbf{Ent-Syn}).

To supervise fine-grained action prediction from human video, each target
action chunk requires a localized visual reference in the demonstration.
When human and robot videos are paired only by task label or synthesized from
an edited initial frame~\cite{zhou2026zerowam}, subsequent motion paths and
contact timing remain unaligned with the ground-truth robot actions.
To provide unambiguous supervision, we enforce \emph{strict pairing}: using
the recorded robot trajectory as an anchor, the paired human video matches
(i) the camera viewpoint and scene layout, (ii) the manipulated objects and
sequence of interaction events (e.g., reaching, grasping, transporting, and
releasing), and (iii) the timing of each event.
This alignment requires the policy to track visual progression instead of
relying on initial-scene or task-identity shortcuts; we ablate strict pairing
in Table~\ref{tab:real_robot_main} and evaluate cross-scene transfer at
deployment in Section~\ref{sec:context_ood_evaluation}.

\subsection{Ent-Real}
\label{sec:real_paired_data}

We construct real pairs by having humans reenact recorded
Astribot S1 trajectories in the same workspace under a viewpoint matched to the
robot's head camera.
During recording, demonstrators watch the source robot video playback and
perform the manipulation in synchrony, matching the object interactions, motion
paths, and execution tempo.

\subsection{Ent-Syn}
\label{sec:synthetic_paired_data}

To scale the paired corpus without per-episode human reenactment, we convert
existing robot videos into human demonstrations through a three-stage pipeline
(Fig.~\ref{fig:paired_dataset_overview}(a)).

\noindent\textbf{Keyframe editing.}
We extract atomic action descriptions from the robot-video annotations
(including Qwen3.8-generated descriptions) and sample source keyframes for
appearance editing.
Using human-hand images from HaGRID~\cite{kapitanov2024hagrid} as appearance references,
\textit{gpt-image-2} replaces visible robot arms and grippers with human hands
and forearms while preserving the scene, task objects, viewpoint, and contact
relationships.
The edited frames pass visual quality checks before video generation.

\noindent\textbf{Video generation.}
Using \textit{Seedance 2.5} or \textit{Wan 3.0}, the video editor takes the complete source robot
clip as a motion reference, the edited keyframes as appearance references, and
a text prompt with the atomic action description.
Supplying the full robot clip anchors action timing, object trajectories, and
state transitions throughout generation without requiring explicit segmentation
masks, depth maps, or edge-map control.

\noindent\textbf{Verification and pairing.}
A vision--language model (VLM) verifier filters generated clips for visual quality (complete arm replacement and scene stability) and temporal action consistency with the source trajectory; verified clips are paired with the original robot observations and actions (details in Appendix~\ref{sec:paired_data_details}).

\section{Methods}
\label{sec:methods}

We first formulate human-video-conditioned in-context manipulation, then
present the \ours architecture and its training recipe combining an intra-task
InfoNCE contrastive objective with variable-length window training.

\subsection{Preliminaries}
\label{sec:formulation}

\noindent\textbf{Problem formulation.}
We consider in-context robotic manipulation conditioned on a single human
demonstration video $\mathcal V^{\mathrm{ctx}}=(v_1^{\mathrm{ctx}},\ldots,v_K^{\mathrm{ctx}})$
that specifies the target manipulation behavior.
Given $\mathcal V^{\mathrm{ctx}}$ and the current robot visual observation $I_t$,
the policy predicts future robot video frames
$V=(I_{t+1},\ldots,I_{t+T_v})$ and an executable action chunk
$A=(a_t,\ldots,a_{t+T_a-1})$ by modeling their conditional joint distribution:
\begin{equation}
\label{eq:icl_objective}
p_\theta\!\left(V,A \mid I_t,\mathcal V^{\mathrm{ctx}}\right).
\end{equation}
Jointly modeling future visual states and actions couples visual dynamics
prediction with control under the context demonstration.
At test time, model parameters $\theta$ remain fixed: the robot executes a
predicted action prefix and replans closed-loop from updated observations using
the same demonstration $\mathcal V^{\mathrm{ctx}}$.

\noindent\textbf{Conditional flow matching.}
We train the video--action backbone with conditional flow
matching~\cite{lipman2022flow}. For each modality $m\in\{v,a\}$,
clean future video latents $X_0^v=\mathcal E(V)$ or normalized actions
$X_0^a$ are interpolated with independent Gaussian noise
$\epsilon^m\sim\mathcal N(0,I)$ as
$X_{\sigma_m}^m=(1-\sigma_m)X_0^m+\sigma_m\epsilon^m$,
where $\sigma_m\in[0,1]$.
Conditioned on the human context and current robot observation, the model
predicts the target velocity $\epsilon^m-X_0^m$ using
\begin{equation}
\begin{aligned}
\mathcal L_m&=\mathbb E\!\left[\|u_\theta^m-(\epsilon^m-X_0^m)\|_{M_m}^2\right],\\
\mathcal L_{\mathrm{WAM}}&=\mathcal L_v+\lambda_a\mathcal L_a,
\end{aligned}
\label{eq:wam_loss}
\end{equation}
where $M_m$ excludes clean conditioning tokens and $\lambda_a$ weights the
action loss. At inference, video and action samples are jointly denoised
from noise and decoded into future frames and executable actions.

\subsection{Model Architecture}
\label{sec:architecture}

\begin{figure}[!htbp]
  \centering
  \includegraphics[width=\textwidth]{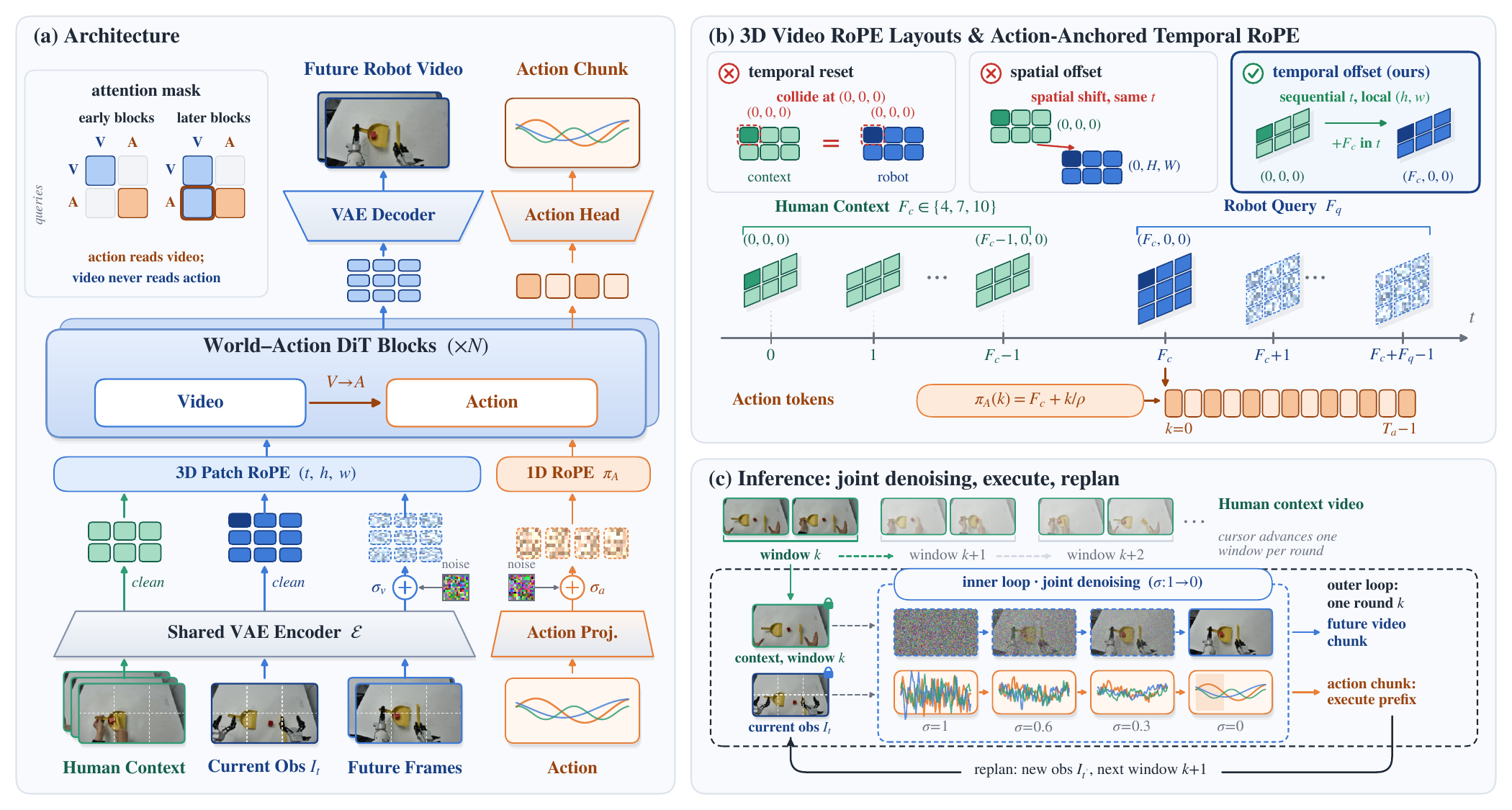}
  \caption{\textbf{Overview of \ours.}
  (a) Human context and the current robot observation condition joint
  video--action prediction, with action tokens attending to video tokens.
  (b) Temporal RoPE places context and robot video sequentially while
  preserving local spatial coordinates, and anchors action positions to
  the robot segment.
  (c) Each inference round jointly denoises future video and actions,
  executes an action prefix, and replans with updated observations and
  context.}
  \label{fig:model_architecture}
\end{figure}

Figure~\ref{fig:model_architecture}(a) illustrates the \ours architecture.
The human demonstration $\mathcal V^{\mathrm{ctx}}$ and the robot video
$(I_t, V)$ are encoded by the shared 3D VAE $\mathcal E$ into spatiotemporal
latent grids of $F_c$ and $F_q$ frames, respectively, where the demonstration
latents and current observation $I_t$ remain clean ($\sigma_v=0$) while future
robot latents $X_{\sigma_v}^v$ are noised.
The patchified video tokens form a unified stream $H_V$, processed alongside
the projected action tokens $H_A$.

\noindent\textbf{Mixture-of-Transformers with directional $V\!\rightarrow\!A$ attention.}
Following a Mixture-of-Transformers (MoT) design~\cite{bi2026motus,zhou2026zerowam,motubrain2026},
our World--Action DiT backbone keeps modality-specific QKV projections,
normalization layers, and feed-forward networks for the video and action
streams, and couples the two only through shared attention.
As in the base model, early blocks run the streams independently; in the
remaining blocks, action queries attend to video keys, whereas video queries
never attend to action keys (Fig.~\ref{fig:model_architecture}(a)).
This asymmetry lets the action stream read context-conditioned visual features
without noisy action tokens perturbing video generation.

\noindent\textbf{Temporal RoPE alignment.}
Video tokens use 3D rotary positional embeddings (RoPE) over temporal, height,
and width coordinates $(t,h,w)$, alongside additive camera-view embeddings;
action tokens use 1D temporal RoPE.
The three layouts in Fig.~\ref{fig:model_architecture}(b) differ only in which
axis is reset or offset per segment: temporal reset shares the spatial grid
$(h,w)$ but assigns both segments identical $t$; spatial offset makes $(h,w)$
disjoint but breaks cross-segment correspondence; our temporal offset makes $t$
disjoint while sharing $(h,w)$. This layout assigns the human context
($t\in\{0,\ldots,F_c-1\}$) and robot video ($t\in\{F_c,\ldots,F_c+F_q-1\}$)
sequentially along the time axis, preserving spatial correspondence without
temporal collision.
Correspondingly, action temporal positions are anchored to the start of the
robot video segment instead of $t=0$ (Fig.~\ref{fig:model_architecture}(b)):
\begin{equation}
\pi_A(k)=F_c+\frac{k}{\rho},
\label{eq:action_rope_anchor}
\end{equation}
where $k$ is the action step index and $\rho$ is the action-to-latent temporal
ratio.
Dynamically shifting $\pi_A(k)$ by $F_c$ keeps action tokens temporally aligned
with the robot video segment across variable context lengths.

\subsection{Training}
\label{sec:training}

We initialize from MotuBrain~\cite{motubrain2026} and continue training (CT) on the full Ent-Dataset (Sec.~\ref{sec:paired_dataset}) with the objectives below. We evaluate CT directly on the real robot; RoboTwin uses multi-task supervised fine-tuning (SFT) for simulator adaptation.

\noindent\textbf{Intra-task InfoNCE contrastive objective.}
The flow-matching loss $\mathcal L_{\mathrm{WAM}}$ (Eq.~\ref{eq:wam_loss})
supervises each trajectory in isolation: when a task admits multiple valid
executions (e.g., different visiting orders or obstacle-avoidance paths), it
does not explicitly penalize predictions that drift toward other valid
trajectories of the same task, allowing the policy to rely on coarse task
identity inferred from $I_t$ without tracking $\mathcal V^{\mathrm{ctx}}$.
Inspired by negative-aware fine-tuning~\cite{zheng2026diffusionnft},
we introduce an intra-task InfoNCE objective~\cite{oord2018representation} that
contrasts the one-step clean-action estimate
$\widehat A_i=X_i-\sigma_i u_{\theta,i}^a$ (from noised action input
$X_i=X_{\sigma_i,i}^a$ at noise level $\sigma_i$) against other trajectories of
the \emph{same} task in the batch.
Let $\mathcal G_i$ denote same-task samples in the batch ($|\mathcal G_i|\ge 2$)
and $d(\widehat A_i, A_j)$ the mean squared distance between $\widehat A_i$ and
clean trajectory $A_j$.
We pull $\widehat A_i$ toward its context-paired trajectory $A_i$ while
repelling same-task alternatives $\{A_j\}_{j\in\mathcal G_i\setminus\{i\}}$ so
task identity alone provides no shortcut:
\begin{equation}
\mathcal L_{\mathrm{InfoNCE}}=\frac{1}{|\mathcal B|}\sum_{i\in\mathcal B}
-\log\frac{\exp[-d(\widehat A_i,A_i)/\tau]}
{\sum_{j\in\mathcal G_i}\exp[-d(\widehat A_i,A_j)/\tau]},
\label{eq:nft}
\end{equation}
where $\tau$ is a temperature and $\mathcal B$ the set of valid anchors, giving
the total loss
$\mathcal{L}=\mathcal{L}_{\mathrm{WAM}}+\lambda_{\mathrm{InfoNCE}}\mathcal{L}_{\mathrm{InfoNCE}}$.

\noindent\textbf{Variable-length window training.}
Context window duration governs a trade-off between procedural following and
generalization: conditioning on full-episode clips encourages the model to
infer global task identity and complete the task semantically instead of
following the context, whereas training only on short clips limits temporal
context and hurts generalization.
To balance this trade-off, we randomly sample synchronized human and robot
windows across multiple durations (13, 25, or 37 frames, encoding to
$F_c\in\{4,7,10\}$ latent frames) with the matching action segment; the
anchored RoPE in Eq.~\ref{eq:action_rope_anchor} keeps actions aligned at every
window length.

\noindent\textbf{Closed-loop inference with Adaptive Horizon (AH).}
When executing a predicted action chunk open-loop, later steps can lose grounding
in the human reference as future uncertainty grows, accumulating contact and
trajectory errors before the next observation.
In each replanning round (Fig.~\ref{fig:model_architecture}(c)), we therefore
monitor $V\!\rightarrow\!A$ attention from predicted actions to the reference
tokens: when later steps exhibit low reference-attention weight and high temporal
entropy (indicating weak, diffuse grounding in the demonstration), Adaptive
Horizon (AH) truncates execution to the reliable prefix (6--24 steps), advances
the context window by the executed duration, and replans from fresh
observations. Appendix~\ref{sec:real_inference_protocol} details AH and context synchronization.

\begin{table}[!htbp]
  \centering\fontsize{8}{10}\selectfont
  \setlength{\tabcolsep}{1.2pt}
  \renewcommand{\arraystretch}{1.06}
  \setlength{\heavyrulewidth}{0.9pt}
  \setlength{\lightrulewidth}{0.45pt}
  \newcommand{\simResult}[2]{\ensuremath{#1{\scriptstyle\,\pm\,#2}}}
  \begin{tabularx}{\textwidth}{ll|*{7}{>{\centering\arraybackslash}X}|>{\centering\arraybackslash}X}
    \toprule
    Group & Method & OS & SS & OM & MP & BB & EC & SB & Overall \\
    \midrule
    \multirow{4}{*}{\textit{Baselines}} & WAN-Action~\cite{zhou2026zerowam} & \simResult{3.00}{2.16} & \simResult{7.33}{1.25} & \simResult{2.26}{1.60} & \simResult{10.67}{1.70} & \simResult{15.26}{2.55} & \simResult{38.33}{2.05} & \simResult{0.00}{0.00} & \simResult{10.98}{1.07} \\
     & LingBot-VA~\cite{li2026causal} & \simResult{6.17}{4.87} & \simResult{3.67}{2.49} & \simResult{\underline{29.33}}{10.66} & \simResult{23.33}{8.22} & \simResult{17.33}{6.18} & \simResult{42.33}{7.85} & \simResult{0.00}{0.00} & \simResult{17.45}{1.40} \\
     & Zero-WAM~\cite{zhou2026zerowam} & \simResult{24.67}{2.05} & \simResult{47.00}{4.55} & \simResult{\mathbf{59.00}}{2.83} & \simResult{69.14}{2.93} & \simResult{35.00}{3.74} & \simResult{\mathbf{84.87}}{0.18} & \simResult{9.00}{2.16} & \simResult{46.95}{0.72} \\
     & RoboICL$^{*}$~\cite{liu2026roboicl} & \simResult{\underline{53.33}}{25.17} & \simResult{66.67}{5.77} & \simResult{3.33}{5.77} & \simResult{70.00}{26.46} & \simResult{20.00}{10.00} & \simResult{13.33}{5.77} & \simResult{0.00}{0.00} & \simResult{32.38}{6.44} \\
    \midrule
    \multirow{3}{*}{\textit{Ablations}} & Temporal reset & \simResult{\mathbf{62.00}}{2.65} & \simResult{\underline{77.67}}{5.03} & \simResult{3.00}{1.73} & \simResult{\underline{82.33}}{4.73} & \simResult{\mathbf{94.00}}{3.00} & \simResult{73.33}{7.09} & \simResult{\underline{61.00}}{1.73} & \simResult{\underline{64.76}}{2.65} \\
     & Spatial offset & \simResult{43.33}{4.62} & \simResult{71.33}{2.31} & \simResult{2.00}{0.00} & \simResult{75.67}{3.06} & \simResult{83.67}{1.53} & \simResult{65.33}{1.15} & \simResult{0.00}{0.00} & \simResult{48.76}{1.25} \\
     & w/o ICL & \simResult{25.33}{2.89} & \simResult{56.33}{8.08} & \simResult{19.67}{4.16} & \simResult{57.67}{3.06} & \simResult{\underline{92.67}}{2.08} & \simResult{69.33}{4.51} & \simResult{2.33}{1.15} & \simResult{46.19}{1.92} \\
    \midrule
    \rowcolor{tableOurs}
    \textit{Ours} & \ours (Ours) & \simResult{46.00}{0.00} & \simResult{\mathbf{91.67}}{3.21} & \simResult{27.33}{4.16} & \simResult{\mathbf{86.67}}{1.53} & \simResult{90.33}{1.53} & \simResult{\underline{74.33}}{2.08} & \simResult{\mathbf{70.33}}{3.79} & \simResult{\mathbf{69.52}}{0.64} \\
    \bottomrule
  \end{tabularx}
  \caption{RoboTwin success (\%, $\uparrow$): mean $\pm$ SD over three seeds; best/second-best means bold/underlined. Tasks: OS, object on scale; SS, stamp seal; OM, open microwave; MP, stapler to pad; BB, bread in basket; EC, empty cup on coaster; SB, stack three blocks. WAN-Action, LingBot-VA, and Zero-WAM results come from Zero-WAM Table~2~\cite{zhou2026zerowam}; $^{*}$RoboICL reports first-attempt success (Appendix~\ref{sec:appendix_inference_settings}).}
  \label{tab:simulation_main}
\end{table}

\section{Experiments}
\label{sec:experiments}

We evaluate \ours on the RoboTwin benchmark~\cite{mu2025robotwin,chen2025robotwin2} and a physical Astribot S1 robot across two generalization axes: \textbf{Task-OOD}, where the complete goal and multi-step procedure are absent from training data, and \textbf{Context-OOD}, where the execution scene departs from strictly aligned training pairs through rearranged layouts or substituted objects and tools.

\subsection{Experimental Setup}
\label{sec:experimental_setup}

\noindent\textbf{Inference protocol.}
Both simulation and real-robot inference condition purely on a 13-frame human-video window without language input, using 50 denoising steps and video/action guidance scales of 5 and 1.
\noindent\textbf{Simulation (RoboTwin).} We address the simulator domain shift with joint SFT on 43 RoboTwin training tasks, then evaluate on seven unseen tasks. Context advances 13 frames per prediction round.
\noindent\textbf{Real robot (Astribot S1).} We evaluate the CT model directly without downstream SFT, predicting 24 action steps at 10\,Hz (2.4\,s) and executing 6--24 steps per round via AH (or all 24 steps without AH; Appendix~\ref{sec:real_inference_protocol}).

\noindent\textbf{Metrics.}
We report binary success rates in simulation and graded task scores (0--100) on the real robot. Real-robot scores are averaged over 10 trials per task and configuration, measuring completion of both the demonstrated goal and procedural constraints (rubrics in Appendix~\ref{sec:real_task_scoring}).

\subsection{RoboTwin Benchmark Evaluation}
\label{sec:simulation_evaluation}

We evaluate seven RoboTwin tasks excluded from SFT training under the Zero-WAM \texttt{demo\_clean} protocol~\cite{zhou2026zerowam}: three seeds and 100 episodes per task per seed, with one fixed generated human context per task. \ours reaches $69.52\%$, exceeding Zero-WAM ($46.95\%$) by 22.57 percentage points (Table~\ref{tab:simulation_main}) and the agent-based RoboICL (32.38\%) by 37.14 points.
While Zero-WAM competes well on single-object tasks (e.g., opening a microwave, placing a cup), it scores 9.00\% on three-block stacking (SB), versus Entangle’s 70.33\%.
This gap indicates that task-matched clips generated from an edited initial frame suffice for short, single-stage actions, whereas frame-aligned correspondence is critical for multi-step sequences.

\subsection{Real-Robot Task-OOD Evaluation}
\label{sec:real_robot_evaluation}

\begin{figure}[!htbp]
  \centering
  \includegraphics[width=\textwidth]{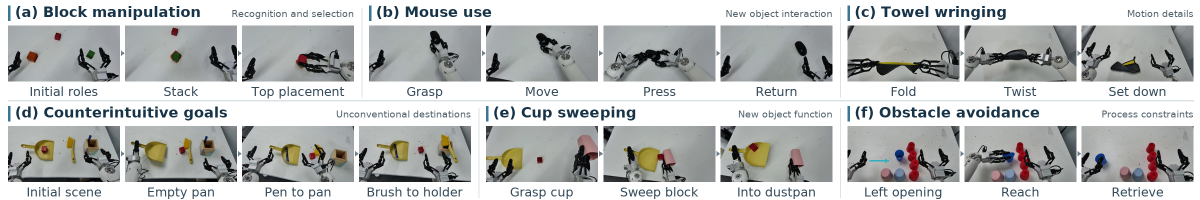}
  \caption{Six real-robot tasks with execution frames in chronological order.
  The cyan arrow in (f) marks the required left opening.}
  \label{fig:real_suite_overview}
\end{figure}

The six Astribot S1 tasks are absent from pretraining and CT. They test \textbf{block manipulation} (object roles and ordering), \textbf{mouse use} (novel grasping, surface motion, and clicks), \textbf{towel wringing} (bimanual coordination), \textbf{counterintuitive goals} (unconventional destinations), \textbf{sweeping} (using a cup as a tool), and \textbf{obstacle avoidance} (following a specified passage beyond reaching the goal; Fig.~\ref{fig:real_suite_overview}). Appendix~\ref{sec:task_ood_details} gives task definitions, and Appendix~\ref{sec:appendix_real_visualizations} pairs human contexts with robot executions.

\paragraph{Component controls.}
Table~\ref{tab:real_robot_main} ablates strict pairing, same-task trajectory contrast (InfoNCE), variable-length training (VL), and adaptive horizon (AH); the full model leads on five of six tasks (83.5 mean). Replacing a reference with another same-task episode at training time (w/o pairing, 41.3) conveys coarse intent on Blocks, but collapses fine bimanual twisting (Towel), trajectory tracking (Mouse), and counterintuitive placement: task-level pairing cannot supervise fine motion or counter-prior goals. Without InfoNCE (64.7), the policy drifts toward an average path that collides with the obstacle, degrading multi-route tasks (Avoidance and Mouse). Fixed 13-frame training (w/o VL, 62.7) restricts the model to myopic local mappings that fail to connect the detour passage to the goal, reducing Avoidance to zero. Finally, executing all 24 steps open-loop (w/o AH, 61.3) hurts Avoidance and Towel, as executing later steps after reference attention weakens accumulates route and contact errors. Thus, data-level alignment, training-level contrast, and deployment-level replanning are all essential.

\begin{table}[t]
  \centering\footnotesize
  \setlength{\tabcolsep}{1.5pt}
  \renewcommand{\arraystretch}{1.15}
  \caption{Astribot S1 ablations: 10-trial mean scores (0--100, $\uparrow$). Best/second-best per column: bold/underlined. Counter.: counterintuitive goals; Avoid.: obstacle avoidance.}
  \label{tab:real_robot_main}
  \begin{tabularx}{\columnwidth}{@{}l*{6}{>{\centering\arraybackslash}X}|r@{}}
    \toprule
    Configuration & Blocks & Mouse & Towel & {\scriptsize Counter.} & Sweep & Avoid. & Avg. \\
    \midrule
    w/o AH & \underline{89} & 71 & 55 & 60 & 65 & 28 & 61.3 \\
    w/o InfoNCE & \underline{89} & 53 & \textbf{83} & \underline{79} & 64 & 20 & \underline{64.7} \\
    w/o VL & 84 & \underline{80} & 71 & 66 & \underline{75} & 0 & 62.7 \\
    w/o pairing & 76 & 26 & 7 & 35 & 66 & \underline{38} & 41.3 \\
    \midrule
    \rowcolor{tableOurs}
    Full & \textbf{92} & \textbf{96} & \underline{76} & \textbf{91} & \textbf{84} & \textbf{62} & \textbf{83.5} \\
    \bottomrule
  \end{tabularx}
\end{table}

\subsection{Ablations}
\label{sec:rope_ablation}

\noindent\textbf{Positional encoding.}
Under the same protocol, our temporal-offset layout reaches $69.52\%$, versus $64.76\%$ for temporal reset and $48.76\%$ for spatial offset (zero on SB; Table~\ref{tab:simulation_main}). The 20.76-point penalty is consistent with shared 2D coordinates across matched viewpoints being important: temporal offsets separate the two segments while preserving that correspondence.

\noindent\textbf{Role of video context.}
A separately trained baseline without human-video context (w/o ICL) reaches $46.19\%$ overall: it is strong when the goal is evident from the initial scene ($92.67\%$ on BB, above the full model's $90.33\%$), but falls to $2.33\%$ on SB, where \ours reaches $70.33\%$. This pattern is consistent with scene priors sufficing for single-stage routines, but not for procedures inferred from the demonstration.

\Needspace{14\baselineskip}
\subsection{Data Scaling on Real-Robot Task-OOD}
\label{sec:data_scaling}

\begin{wrapfigure}{r}{.43\linewidth}
  \centering
  \includegraphics[width=\linewidth]{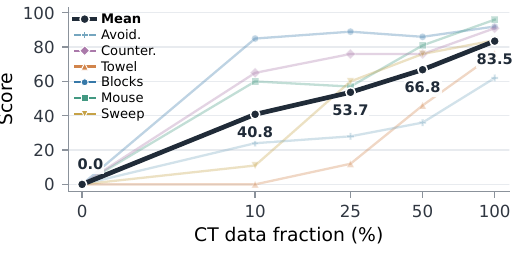}
  \captionsetup{justification=raggedright,singlelinecheck=false}
  \caption{CT-data scaling on six real-robot tasks (10 trials each). Faint curves: task scores; dark: unweighted mean. Positive data fractions are log-spaced.}
  \label{fig:data_scaling}
\end{wrapfigure}

We evaluate five nested CT-data fractions (0\%, 10\%, 25\%, 50\%, and 100\%) on the six real-robot tasks (Fig.~\ref{fig:data_scaling}), where 0\% evaluates the pretrained backbone directly and nonzero fractions train for 4,000 CT steps. Without aligned CT data (0\%), the pretrained backbone scores zero across all tasks; as aligned data scales from 10\% to 100\%, the mean score rises monotonically from 40.8 to 83.5. Across tasks, a small fraction (10\%--25\%) already suffices to ground high-level intent such as object ordering (Blocks), counterintuitive destinations (Counterintuitive), and tool repurposing (Sweeping), whereas continuous bimanual coordination (Towel) and strict route adherence (Avoidance) require 50\%--100\% of the corpus. This indicates that while modest paired data teaches the policy \emph{what} object or goal to target, scaling temporally aligned data via our synthetic pipeline is essential for translating \emph{how} a human moves into fine-grained robot control.
\subsection{Context-OOD Evaluation}
\label{sec:context_ood_evaluation}

\begin{figure}[!htbp]
  \centering
  \includegraphics[width=\textwidth]{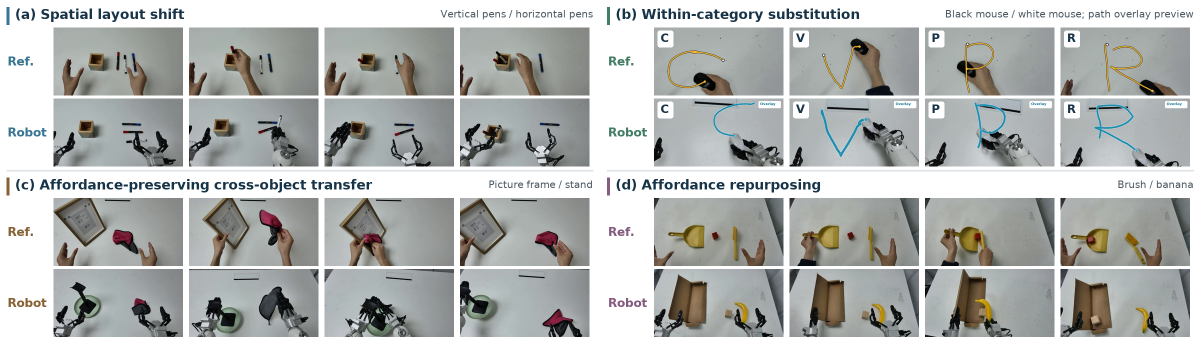}
  \caption{Four types of Context-OOD, shown as compact process sequences. (a) Pen placement with changed spatial layout;
  selected frames show the initial scene and the first two placements.
  (b) Black-to-white mouse substitution: lines denote mouse
  writing strokes.
  (c) Picture-frame-to-stand transfer preserves the hold-and-wipe
  interaction. (d) A banana takes the unconventional role of a sweeping
  tool.}
  \label{fig:context_ood_cases}
\end{figure}

Because all training pairs are strictly scene-aligned, we evaluate four progressively harder Context-OOD shifts (Fig.~\ref{fig:context_ood_cases}) to examine how \ours establishes correspondence when the workspace departs from the reference. Under \textbf{spatial layout shift} (reoriented pens) and \textbf{within-category substitution} (black to white mouse), the policy adapts its grasp orientation and reproduces the same letter-tracing trajectories across mouse appearances. Under \textbf{affordance-preserving cross-object transfer} and \textbf{affordance repurposing}, the policy transfers interactions across object categories, adapting hold-and-wipe from a picture frame to a stand and using a banana to sweep a wooden block into a box when a brush and dustpan are absent. Thus, supervising strict temporal alignment on a world-action prior does not lock the policy into pixel-level copying; instead, \ours exhibits an emergent ability to transfer demonstrated interactions to novel objects and scenes, freeing deployment from strict scene alignment to accommodate diverse demonstrations.

\subsection{Qualitative Analysis}
\label{sec:qualitative_results}

\subsubsection{With and Without InfoNCE}
\label{sec:qualitative_infonce}

\begin{figure}[!htbp]
  \centering
  \includegraphics[width=.90\textwidth]{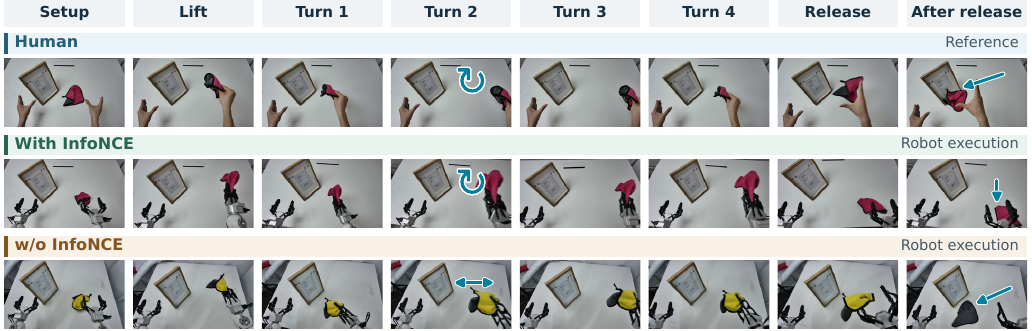}
  \caption{Cloth rotation and toss in two rollouts conditioned on the same human video. Schematic arrows show circular or side-to-side motion during the turning phase and the toss or drop direction in the final column. The cloth color differs across the rollouts.}
  \label{fig:infonce_cloth_rotation}
\end{figure}

\noindent\textbf{Cloth rotation and toss.} Both policies receive the same reference, in which the human lifts a cloth, moves it in a circle, and tosses it toward the picture frame. With InfoNCE, the wrist rotation more closely matches the demonstration. The robot opens its gripper but fails to toss the cloth. Without InfoNCE, the robot only partially follows the demonstrated circular motion but successfully tosses the cloth (Fig.~\ref{fig:infonce_cloth_rotation}).

\begin{figure}[!htbp]
  \centering
  \includegraphics[width=.82\textwidth]{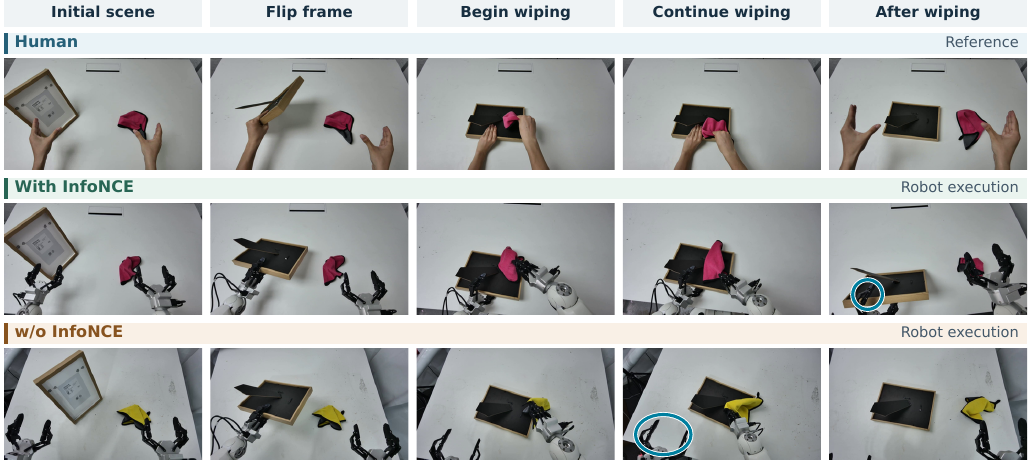}
  \caption{Circles mark the frame snag (middle row) and early left-gripper withdrawal (bottom row).}
  \label{fig:infonce_back_surface_wiping}
\end{figure}

\noindent\textbf{Wiping the back of a picture frame.} Both policies receive the same reference, in which the human flips the frame, steadies it with the left hand while wiping with the right, and withdraws both hands. With InfoNCE, the left gripper holds the frame throughout wiping. The robot then retracts its arms as shown in the context video. During retraction, the frame gets snagged on one of the grippers. Without InfoNCE, the left gripper withdraws early and the right wipes near the edge without the demonstrated support (Fig.~\ref{fig:infonce_back_surface_wiping}).

Intra-task InfoNCE favors the trajectory paired with the current video among alternatives for the same task. In these examples, the policy trained with InfoNCE follows the demonstrated procedure more closely. The runs without it show both the flexibility and the cost of taking alternative trajectories. A different wrist motion produces a successful toss, while the early withdrawal during wiping omits the demonstrated support.

\subsubsection{Deployment with Generated Video Context}
\label{sec:generated_context_evaluation}

\begin{figure}[!htbp]
  \centering
  \includegraphics[width=.60\linewidth]{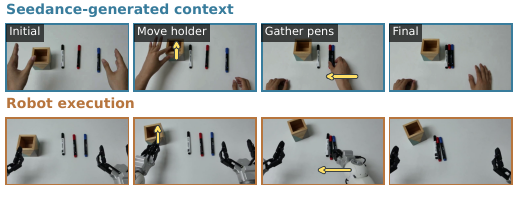}
  \caption{Deployment from Seedance-generated context without a recorded
  human demonstration. Astribot S1 first pushes the pen holder, then gathers
  three pens. Yellow arrows indicate motion.}
  \label{fig:generated_context_case}
\end{figure}

We condition Astribot S1 on a Seedance-generated video. The video shows the left hand pushing the pen holder forward and the right hand gathering three pens beside it. The robot successfully follows the two-step procedure zero-shot (Fig.~\ref{fig:generated_context_case}).

This example illustrates how a video generation model can serve as a zero-shot visual planner~\cite{bao2024vidu,pmlr-v305-bharadhwaj25a}. The generator produces a visual plan, and the policy grounds that plan in the robot's workspace, allowing task specification without recording a human demonstration. It could scale real-robot data collection by enabling robots to autonomously execute a diverse range of tasks beyond those seen during policy training.

\subsubsection{Execution Without Task Objects}
\label{sec:pantomime_execution}

\begin{figure}[!htbp]
  \centering
  \includegraphics[width=.82\textwidth]{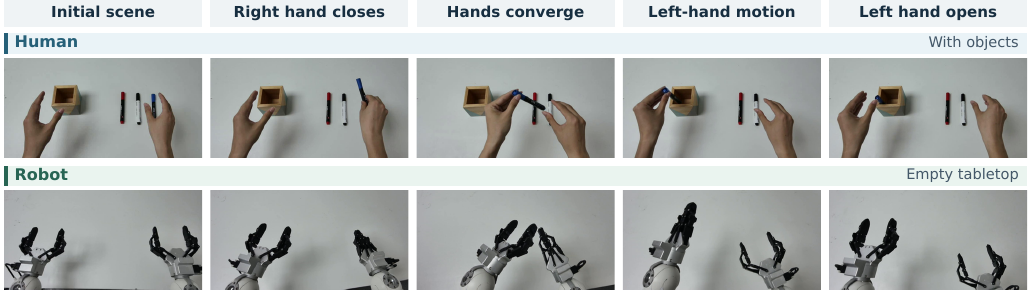}
  \caption{The robot reproduces the demonstrated handover motions on an empty tabletop.}
  \label{fig:pantomime_handover}
\end{figure}

The human picks up a pen with the right hand, passes it to the left hand, and places it in a holder. On an empty tabletop, the robot closes its right gripper, brings both grippers together, and opens the right gripper while the left remains closed. It then opens the left gripper at the end of the placement motion, preserving the order and hand roles of the demonstration (Fig.~\ref{fig:pantomime_handover}).

With the pen and holder absent, the scene provides no object-state cues for the transfer. The preserved handover sequence and release order demonstrate the policy's ability to follow the context video.

\section{Limitations and Future Work}

Our real-robot evaluation is limited to one embodiment, and the data pipeline
relies on the fidelity of generated human videos. Cross-scene transfer and
generated-context deployment are demonstrated through qualitative cases;
broader quantitative evaluation is needed. Future work will explore
cross-embodiment transfer, more reliable data generation, and failure recovery
for longer tasks.

\section{Conclusion}

We presented \ours, which uses human video as an interface for specifying
robot behavior without task-specific parameter updates. Our findings suggest
that learning from aligned human--robot experience can support transferable
demonstration following beyond familiar tasks and scenes. This offers a path
toward more accessible robot adaptation, where people can communicate new
behaviors by showing how they should be performed.

\FloatBarrier
\bibliographystyle{unsrtnat}
\bibliography{main}

\clearpage
\raggedbottom
\beginappendix
\section{Real-Robot Tasks, Environments, and Scoring}
\label{sec:task_ood_details}

\subsection{Capability Coverage and Task Selection}
\label{sec:real_capability_design}
We checked the MotuBrain pretraining and continued training (CT) data for the goal and full procedure of each evaluation task.
None of the six complete tasks appears in either training dataset.

Table~\ref{tab:real_capability_design} identifies the main capability that
motivates each task. All six tasks require multiple operations and can draw
on several capabilities.

\noindent\begin{minipage}{\linewidth}
\centering\small
\setlength{\tabcolsep}{5pt}
\renewcommand{\arraystretch}{1.15}
\captionof{table}{Capabilities emphasized by the six real-robot tasks.}
\label{tab:real_capability_design}
\begin{tabular}{p{.21\textwidth}p{.18\textwidth}p{.51\textwidth}}
\toprule
Capability & Task & What the demonstration specifies \\
\midrule
Recognition and selection & Block manipulation & Select spatial object
roles and follow the left/right/middle placement order. \\
New object interaction & Mouse use & Follow demonstrated grasping,
surface motion, and button presses when interacting with the mouse. \\
Motion-detail acquisition & Towel wringing & Reproduce the demonstrated
folding and coordinated two-handed twisting motions. \\
Counterintuitive goals & Object rearrangement & Follow the unusual
destinations and operation order. \\
New object function & Cup sweeping & Assign the cup a sweeping-tool role
and use it to move the block into the dustpan. \\
Process constraints & Obstacle avoidance & Use the specified opening and
avoid obstacle contact while acquiring and placing the cup. \\
\bottomrule
\end{tabular}
\end{minipage}\par

\subsection{Task Definitions}
\label{sec:real_task_definitions}

\noindent\textbf{Block manipulation.}
Stack three cubes on the table by placing the initially left cube at the base, the initially right cube on it, and the initially middle cube on top.

\noindent\textbf{Mouse use.}
Grasp the mouse, trace C, V, P, and R on the tabletop, press the left button five times, and return the mouse to its starting region.

\noindent\textbf{Towel wringing.}
Lift and fold the towel, hold opposite portions with the two grippers, twist in both directions, and place the towel back on the table.

\noindent\textbf{Counterintuitive goals.}
Dump the block out of the dustpan, move the pen from its holder into the dustpan, and place the brush on top of the pen holder, in that order.
Then return the arms to their resting configuration.
Object positions may vary slightly between trials.

\noindent\textbf{Cup sweeping.}
Grasp the cup and dustpan, sweep the block into the dustpan with the cup, raise the dustpan, and return both tools to their starting regions.

\noindent\textbf{Obstacle avoidance.}
Reach the target cup through the left-side opening, grasp it, and place it stably while avoiding contact with obstacles.
Obstacle identities and positions may vary between trials.

\subsection{Task-Specific Scoring Rubrics}
\label{sec:real_task_scoring}

Start each trial at 100 points, apply the deductions in Table~\ref{tab:real_task_rubric}, and floor the score at zero.
Deductions marked ``Each'' apply separately to every item or event specified in the table (e.g., 10 points per missing or unrecognizable letter).
All other deductions apply at most once per trial.
A trial receives zero if the robot makes no progress on the task.

{\small
\setlength{\tabcolsep}{6pt}
\setlength{\LTpre}{\smallskipamount}
\setlength{\LTpost}{\smallskipamount}
\setlength{\LTcapwidth}{\linewidth}
\renewcommand{\arraystretch}{1.08}
\begin{longtable}{@{}>{\raggedright\arraybackslash}p{.20\linewidth}
  >{\raggedright\arraybackslash}p{.63\linewidth}
  >{\raggedleft\arraybackslash}p{\dimexpr.17\linewidth-4\tabcolsep\relax}@{}}
\caption{Task-specific deductions for real-robot evaluation.}\label{tab:real_task_rubric}\\
\toprule
Task & When to deduct & Points deducted \\
\midrule
\endfirsthead
\toprule
Task & When to deduct & Points deducted \\
\midrule
\endhead
\bottomrule
\endfoot
\textbf{Block manipulation} & Each cube missing from the required stack & 10 \\*
& Incorrect left--right--middle selection order, counted once & 10 \\
\specialrule{.35pt}{1.4pt}{1.4pt}
\textbf{Mouse use} & Each missing or unrecognizable letter (C, V, P, or R) & 10 \\*
& Each letter during which the mouse is lifted more than 5\,cm & 5 \\*
& Each letter during which the mouse is dropped and regrasped & 5 \\*
& Each omitted press out of five & 10 \\*
& Presses that miss the left button, counted once & 5 \\*
& Failure to return the mouse & 5 \\
\specialrule{.35pt}{1.4pt}{1.4pt}
\textbf{Towel wringing} & Missing fold & 10 \\*
& Each loss of grasp & 5 \\*
& Each omitted twist direction & 10 \\*
& Improper final placement & 5 \\
\specialrule{.35pt}{1.4pt}{1.4pt}
\textbf{Counterintuitive goals} & Each failed rearrangement (dumping the block, placing the pen in the pan, or placing the brush on the holder) & 25 \\*
& Failure to return the arms to rest & 20 \\*
& Each completed rearrangement with a minor placement error & 10 \\
\specialrule{.35pt}{1.4pt}{1.4pt}
\textbf{Cup sweeping} & Failure to secure the tools, counted once & 10 \\*
& Block not transferred into the dustpan, counted once & 10 \\*
& Upward dustpan motion omitted & 10 \\*
& Each tool not returned to its starting region & 10 \\
\specialrule{.35pt}{1.4pt}{1.4pt}
\textbf{Obstacle avoidance} & Each distinct obstacle contacted & 20 \\*
& Failure to grasp the target & 20 \\*
& Failure to place the target stably & 20 \\*
& Route other than the prescribed left-side passage & 30 \\
\end{longtable}
}

\section{Additional Experimental Controls}
\label{sec:appendix_additional_results}

\noindent\textbf{Real robot.}
The Adaptive Horizon ablation (\textbf{w/o AH}) uses the same CT checkpoint and executes all 24 predicted actions before replanning (Section~\ref{sec:real_inference_protocol}).
For the InfoNCE ablation (\textbf{w/o InfoNCE}), we omit the contrastive loss term during training.
The fixed-length training control (\textbf{w/o VL}) uses 13-frame windows in place of the mixed 13-, 25-, and 37-frame windows.
To ablate strict pairing (\textbf{w/o pairing}), we replace matched references with clips from different episodes of the same task, retaining the original target trajectories and action supervision.
We fix the replacement assignments before training.
For data scaling, the 0\% baseline uses MotuBrain directly.
The 10\%, 25\%, 50\%, and 100\% training sets are nested and share the same MotuBrain initialization, global batch size of 1,024, optimizer settings, and 4,000-step CT budget.

\noindent\begin{minipage}[t]{.52\linewidth}
\noindent\textbf{RoboTwin.}
Each RoPE variant uses its assigned positional layout during training and inference.
The no-reference model starts from a reference-conditioned CT checkpoint at step 4,000 and undergoes 4,000 RoboTwin SFT steps without human-video context.
We also omit the reference at evaluation.
Table~\ref{tab:robotwin_controls} reports results for three evaluation seeds.

\end{minipage}\hfill%
\begin{minipage}[t]{.45\linewidth}
  \centering\footnotesize
  \setlength{\tabcolsep}{2pt}
  \captionof{table}{RoboTwin control results over three evaluation seeds. Each seed runs 700 episodes with a fixed checkpoint per row. We report success counts and the mean $\pm$ sample standard deviation of the seed-level success rates.}
  \label{tab:robotwin_controls}
  \begin{tabular}{lrrrc}
    \toprule
    Control & Seed 0 & Seed 1 & Seed 2 & Rate (\%) \\
    \midrule
    Temporal offset & 491 & 487 & 482 & $69.52\pm0.64$ \\
    Temporal reset & 432 & 462 & 466 & $64.76\pm2.65$ \\
    Spatial offset & 351 & 334 & 339 & $48.76\pm1.25$ \\
    \midrule
    No reference & 308 & 333 & 329 & $46.19\pm1.92$ \\
    \bottomrule
  \end{tabular}

\end{minipage}\par

\section{Real-Robot Context and Execution Visualizations}
\label{sec:appendix_real_visualizations}
\begin{figure}[!htbp]
  \centering
  \includegraphics[width=.94\linewidth]{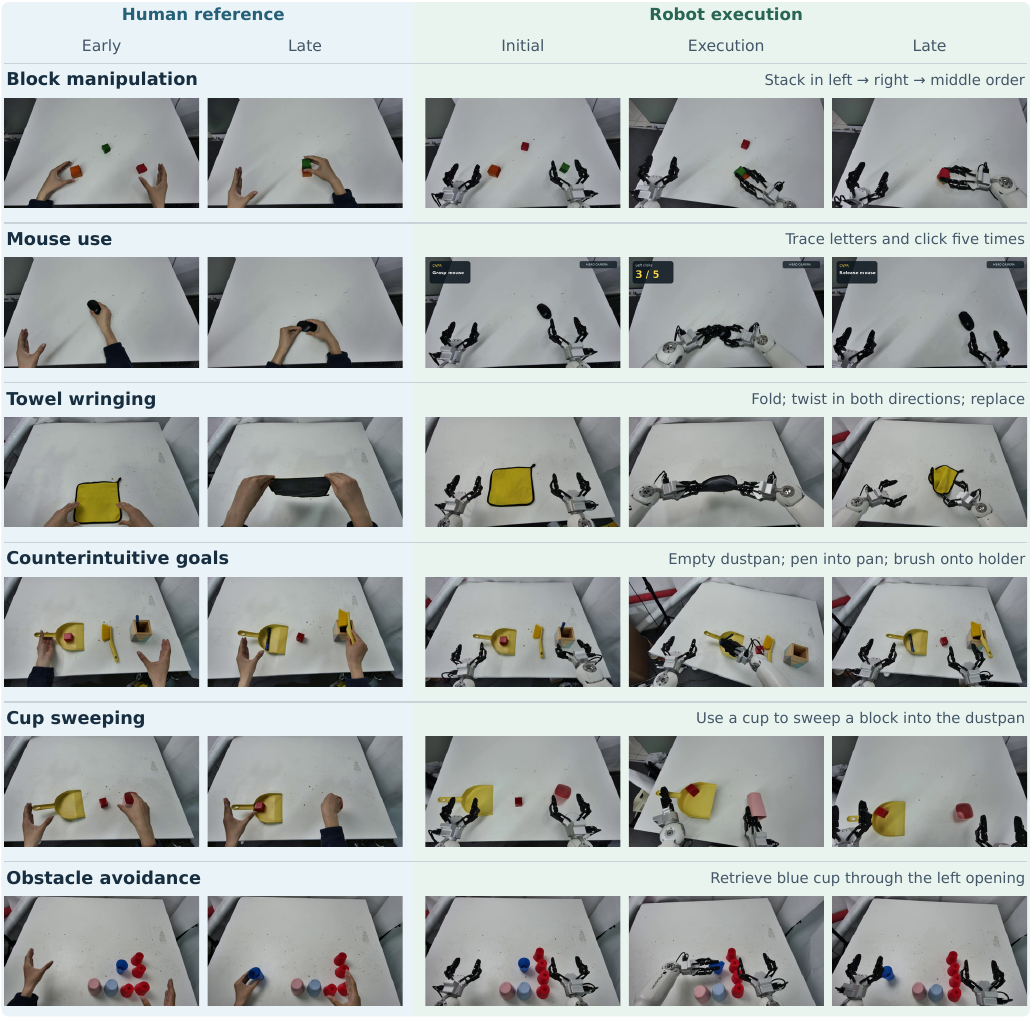}\par
  \caption{Human demonstrations and robot executions for the six real-robot tasks.
  Each row shows early and late demonstration frames, followed by initial, intermediate, and late frames from the robot head camera.}
  \label{fig:real_robot_cases}
\end{figure}

Figure~\ref{fig:real_robot_cases} pairs human demonstrations with execution frames from the robot head camera for each of the six tasks.
The robot head and wrist camera recordings have resolutions of $1280\times720$ and $640\times360$, respectively, at 10\,fps.
We score full episodes using the rubric in Section~\ref{sec:real_task_scoring}.

\section{RoboTwin Tasks, Environments, and Scoring}
\label{sec:appendix_robotwin}

\subsection{Task Environments, Goals, and Evaluation Protocol}
We follow the released Zero-WAM \texttt{demo\_clean} protocol on seven
RoboTwin tasks held out from the 43-task simulation SFT split. Each task
uses one fixed generated human-video reference.
For each video-conditioned checkpoint, we evaluate 100 expert-feasible
scenes per task per seed. A scene qualifies if the scripted expert can
plan and complete the task. We report the mean and sample standard
deviation over three evaluation seeds, averaging the seven task rates
within each seed for the overall score.

\noindent\textbf{Task environments and goals.}
The task definitions and scene setup are identical to those of the original
RoboTwin benchmark.

\noindent\textbf{OS: Object on scale} \textnormal{(place\_object\_scale).}
A mouse, stapler, or bell and an electronic scale are placed on the same
side of the table. The object's initial center has
$x\in[-0.25,0.25]$\,m, $y\in[-0.20,0.05]$\,m, with a small central
exclusion region; the scale is at least 0.15\,m away in the horizontal plane.
Object and scale asset variants and in-plane orientation vary.
Grasp the object, move it onto the scale's platform, and release it.
The score tests placement, not a numerical weight reading.

\noindent\textbf{SS: Stamp seal} \textnormal{(stamp\_seal).}
A seal and a $7\times7$\,cm colored target appear on the same side of the
table, initially separated by at least 0.10\,m. Seal appearance and target
color vary. Move the seal to the designated colored target and release it.
Although the task is called stamping, the implemented predicate checks
horizontal alignment and open grippers; it does not independently verify
stamping pressure or an ink impression.

\noindent\textbf{OM: Open microwave} \textnormal{(open\_microwave).}
A fixed-base microwave with a hinged door is placed toward the back of the
table, with model and position variation. Its base is sampled from
$x\in[-0.12,-0.02]$\,m and $y\in[0.15,0.20]$\,m.
Grasp/pull the handle until the door reaches the required opening angle.
No object insertion, button press, or microwave operation is required.

\noindent\textbf{MP: Move stapler to pad} \textnormal{(move\_stapler\_pad).}
A stapler and an $11\times6$\,cm colored pad are placed on the same side
of the table, initially at least 0.10\,m apart. Stapler model, yaw, and pad
color vary. Grasp the stapler, transfer it to the pad, and release it.
The released Zero-WAM rule requires physical stapler--pad contact with a
horizontal positional margin; it does not impose the upstream task's
specific quaternion condition.

\noindent\textbf{BB: Bread in basket} \textnormal{(place\_bread\_basket).}
A basket is placed near the table center at $(0,-0.20)$\,m. Bread pieces
are sampled in the side regions, separated from the basket and from one
another, with appearance and orientation variation. The scene constructor
attempts to place two pieces; the predicate applies to every piece actually
created. Transfer all pieces into the basket and release them. Leaving one
piece outside does not satisfy the task.

\noindent\textbf{EC: Cup on coaster} \textnormal{(place\_empty\_cup).}
An empty cup begins in a left or right side region, and a fixed coaster is
placed closer to the center, initially at least 0.10\,m away. Grasp the cup,
move it onto the coaster, and release it. The score compares functional
points on the cup and coaster; it does not separately test cup orientation.

\noindent\textbf{SB: Stack three blocks} \textnormal{(stack\_blocks\_three).}
Three 5\,cm cubes, colored red, green, and blue, begin at separated table
positions with varied in-plane rotations. Form a three-block tower with
red at the bottom, green in the middle, and blue on top, then release both
grippers. The native predicate tests relative block alignment and vertical
spacing; it does not require the tower to be at a particular table-center
coordinate.

\subsection{Success Criteria}
\label{sec:robotwin_task_scoring}
All seven tasks use the binary success criteria from the released
Zero-WAM evaluation (Table~\ref{tab:robotwin_task_rubric}). Zero-WAM
provides its own checks for SS and MP and uses RoboTwin's native checks
for the other five tasks. Lengths are in meters. The quantities
$\Delta x$, $\Delta y$, and $\Delta z$ denote coordinate differences.
\emph{Both open} refers to the simulator's left and right gripper open checks.

\noindent\begin{minipage}{\linewidth}
{\centering\small
\setlength{\tabcolsep}{5pt}
\renewcommand{\arraystretch}{1.08}
\captionof{table}{Binary success criteria from the released Zero-WAM
evaluation. All conditions in a row must hold.}
\label{tab:robotwin_task_rubric}
\begin{tabular}{@{}l>{\raggedright\arraybackslash}p{.91\linewidth}@{}}
\toprule
Task & Conditions for success \\
\midrule
OS & The horizontal distance from the object center to the scale's
functional point is $<0.035$. The object height is $>z_{\rm scale}-0.01$,
and the manipulating gripper is open. \\
SS & Seal-to-target $|\Delta x|<0.03$ and $|\Delta y|<0.03$, with both open. \\
OM & Door joint coordinate $q\geq0.6q_{\rm upper}$, where $q_{\rm upper}$
is the joint's upper limit. \\
MP & Stapler-to-pad $|\Delta x|<0.080$ and $|\Delta y|<0.055$, with
physical stapler--pad contact and both open. \\
BB & Every bread piece satisfies basket-relative $|\Delta x|<0.05$,
$|\Delta y|<0.05$, and $z>0.73+\mathrm{table\_z\_bias}$.
Both grippers are open. \\
EC & Cup-to-coaster functional-point horizontal distance $<0.035$ and
$|\Delta z|<0.015$, with both open. \\
SB & For green relative to red and blue relative to green,
$|\Delta x|<0.025$, $|\Delta y|<0.025$, and $|\Delta z-0.05|<0.012$.
Both grippers are open. \\
\bottomrule
\end{tabular}
\par}
\end{minipage}\par

\subsection{Context and Recorded Execution for Every Task}
Figures~\ref{fig:appendix_robotwin_cases} and
\ref{fig:appendix_robotwin_cases_2} show the fixed human contexts and
recorded robot observations for all seven tasks. The robot panels use
one successful rollout per task from the temporal-reset model after
4,000 SFT steps.
The reference videos are the fixed generated clips from the released
evaluation.

\nopagebreak[3]
\noindent\begin{minipage}{\linewidth}
  \centering
  \includegraphics[page=1,width=.96\linewidth]{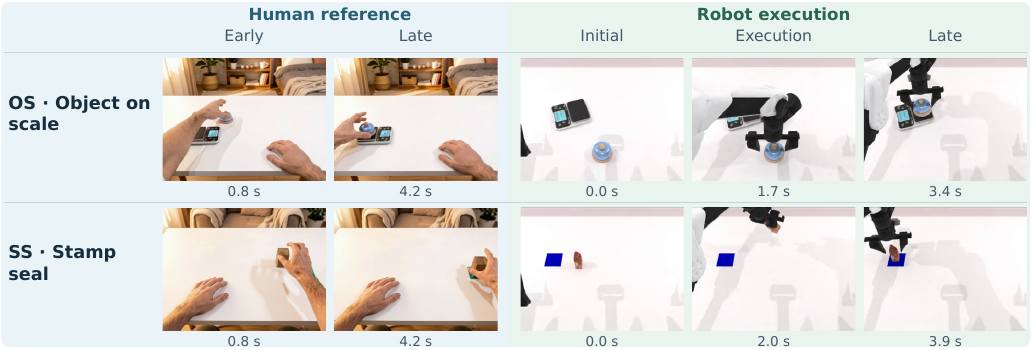}\par
  \captionof*{figure}{\figurename~\ref{fig:appendix_robotwin_cases} (part 1 of 2).}
\end{minipage}\par
\noindent\begin{minipage}{\linewidth}
  \centering
  \includegraphics[page=2,width=.96\linewidth]{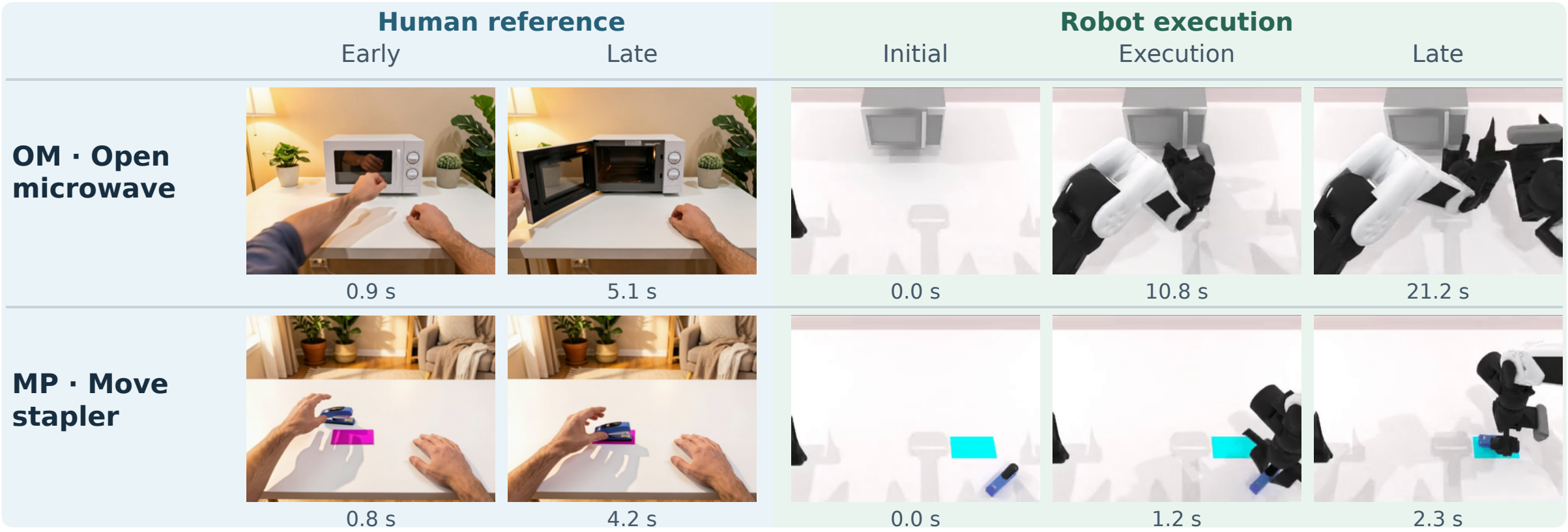}\par
  \captionof{figure}{RoboTwin task environments and interactions for OS, SS, OM, and
  MP. Two human-context keyframes precede initial, intermediate, and late
  robot observations from the temporal-reset model after 4,000 SFT steps.
  Robot panels show the head-camera view.
  Timestamps are relative to each source video.}
  \label{fig:appendix_robotwin_cases}
\end{minipage}\par
\noindent\begin{minipage}{\linewidth}
  \centering
  \includegraphics[width=.96\linewidth]{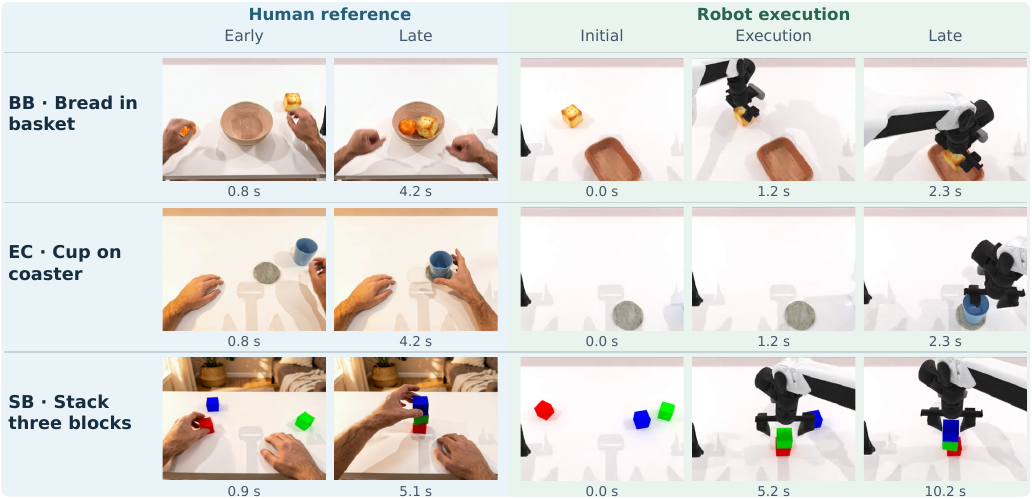}\par
  \captionof{figure}{RoboTwin task environments and interactions for BB, EC, and SB,
  using the same layout and temporal-reset model as
  Figure~\ref{fig:appendix_robotwin_cases}.}
  \label{fig:appendix_robotwin_cases_2}
\end{minipage}\par

\section{Implementation and Evaluation Details}
\label{sec:appendix_implementation}

\subsection{Training Configurations and Checkpoint Lineage}
Real-robot evaluation uses CT checkpoints without downstream real-robot SFT.
The zero-CT condition uses the original MotuBrain EMA.
The nonzero scaling conditions use CT step-4,000 EMA checkpoints.
The scaling subsets are drawn from source-stratified blocks of two
same-task examples and share the full-data validation split.
The simulation branch applies multi-task RoboTwin SFT after CT to adapt
the model to the simulator.

\begin{table}[!htbp]
\centering\small
\setlength{\tabcolsep}{5pt}
\renewcommand{\arraystretch}{1.15}
\caption{Training settings for the real-robot scaling and RoboTwin experiments.
The CT step budget refers to the scaling runs. The RoboTwin SFT models
initialize from the corresponding CT EMA weights.}
\label{tab:appendix_training_settings}
\begin{tabular}{p{.28\textwidth}p{.30\textwidth}p{.31\textwidth}}
\toprule
Setting & CT / real-robot branch & RoboTwin SFT branch \\
\midrule
Training stages & MotuBrain $\rightarrow$ CT & MotuBrain $\rightarrow$ CT $\rightarrow$ SFT \\
Evaluated weights & CT4,000 EMA; MotuBrain EMA for 0\% & Corresponding SFT4,000 EMA \\
Stage length & 4,000 CT steps for nonzero fractions & 4,000 SFT steps \\
Global batch size & 1,024 & 128 \\
SFT optimizer and schedule & --- & AdamW; learning rate $2\times10^{-5}$; 100 warmup steps, then constant \\
SFT microbatch / seed & --- & 2 / initialization seed 6198 \\
Mixed training windows & 13/25/37 video frames in the full recipe & 13/25/37 video frames \\
InfoNCE coefficient / temperature & 0.05 / 0.1 in the InfoNCE recipe & 0.05 / 0.1 \\
Action statistics & CT-specific statistics; real-robot action mapping & RoboTwin training-only, stepwise q99 \\
Evaluation caption & Empty & Empty \\
Checkpoint selection & Saved terminal EMA & Saved terminal EMA \\
\bottomrule
\end{tabular}
\end{table}

\noindent\textbf{Model configuration.}
The model configuration has 24 transformer layers, hidden width 4,096,
32 attention heads, and a 55-dimensional unified action representation.
It uses additive embeddings for the head, left-wrist, and right-wrist views.
The three RoPE conditions use the same layout at training and inference:
temporal offset, temporal reset, and spatial offset.

\subsection{Inference Settings}
\label{sec:appendix_inference_settings}
\noindent\textbf{Simulation.}
The simulation evaluation client combines the head and wrist views into a
$384\times464$ image and uses current-observation geometry features.
At each action inference, the context contains 13 consecutive raw frames,
advances by 13 frames after a successful inference, and repeats its last
frame once the context is exhausted. The context cursor resets per episode;
it does not use the future query trajectory or a learned alignment signal.
The client configures 50 video and 50 action denoising steps, video/action
shifts of 5/1, and context/action guidance of 5/1 with text guidance disabled.
The released controller executes 16 commands initially and 32 subsequently,
interpolating from the source action rate of 10\,Hz to simulator commands at
50\,Hz.

The simulation action mapping uses camera-frame end-effector representations
with 6D rotations and stepwise q99 statistics from RoboTwin training data.
Its gripper channel retains the raw $[0,100]$ convention.
The no-reference control uses a global batch size of 1,024 during SFT.
At inference it receives zero context tokens and uses a context guidance scale of 1.

\noindent\textbf{Real robot.}
Recording resolutions and frame rates for the head and two wrist RGB views
are given in
Section~\ref{sec:appendix_real_visualizations}. Each inference predicts
24 actions at 10\,Hz (2.4\,s); AH selects a 6--24-step execution prefix,
while the no-AH control executes all 24 steps. Context windows contain 13 frames
at 5\,Hz and advance by the executed prefix duration.
Section~\ref{sec:real_inference_protocol} specifies synchronization,
attention extraction, and guidance.
The 10\,Hz rate refers to action waypoints.

\noindent\textbf{External baselines.}
The main-paper WAN-Action, LingBot-VA, and Zero-WAM entries reproduce
measurements from Table~2 of the Zero-WAM paper cited in the main text.
The source uses three evaluation seeds and 100 rollouts per
unseen task per seed, reporting the across-seed mean and standard deviation.
Our evaluation follows the same seven-task split and released success
checks. The results report means and standard deviations over three
evaluation seeds for each fixed checkpoint.

\noindent\textbf{RoboICL evaluation.}
We evaluate the GPT-6 Astra agent with \texttt{xhigh} reasoning effort.
We use the RoboTwin task settings and success predicates described in Sections~\ref{sec:appendix_robotwin} and~\ref{sec:robotwin_task_scoring}.
RoboICL retains interaction history within an episode and can retry failed operations.
For fairness, the starred entry counts success only when every required operation succeeds on its first physical attempt.
We evaluate seven tasks over three seeds, with 10 scenes per task per seed.

\subsection{Real-Robot Context Synchronization and Adaptive Execution}
\label{sec:real_inference_protocol}

\noindent\textbf{Context construction and advancement.}
The full human demonstration is available as a reference, but each model
call receives a local window. At round $k$, with context start time
$\tau_k$, we sample the source video at
$\tau_k+\{0,0.2,\ldots,2.4\}$ seconds. This gives 13 frames at 5\,Hz,
spanning 2.4\,s from first to last frame. The nearest source frame is
selected using its timestamp, so playback time does not depend on the
source video's native frame rate. Each episode starts at $\tau_0=0$.
For an executed prefix of $h_k$ actions at 10\,Hz, we update
\begin{equation}
  \tau_{k+1}=\tau_k+\frac{h_k}{10}.
  \label{eq:real_context_advance}
\end{equation}
The reference-speed multiplier is 1. Inference, network, and waiting
latency do not advance the context clock. For example, executing 10
actions after observing context $[0,2.4]$\,s yields the next window
$[1.0,3.4]$\,s; executing all 24 actions yields $[2.4,4.8]$\,s.
Short executions therefore retain a full context window while creating
overlap between successive windows. This alignment uses cumulative issued
action steps under a synchronous executor that completes each prefix
before requesting the next observation. It does not estimate task progress
from encoders or learn a context--robot phase alignment. Samples beyond
the video end repeat its last frame; the demonstration is not looped.
Starting a new episode or switching the reference resets the context clock.

\noindent\textbf{Prediction, execution, and replanning.}
Every call receives the latest head/left-wrist/right-wrist RGB images,
robot state, and the local human context. The model always predicts 24
future actions at 10\,Hz, representing 2.4\,s of action time.
Only the first robot-video latent time position is conditioned on the
current observation; future positions are initialized from noise, so no
true future robot frames are provided. AH selects an execution prefix of
6--24 actions. After that prefix completes, the unused suffix is discarded
and a fresh 24-action prediction is generated from new observations.
Without AH, every call executes the complete 24-action prediction.
The 10\,Hz rate describes model action-waypoint timing; a lower-level
executor may interpolate at a higher rate. The 0.6--2.4\,s prefix durations
exclude inference latency and are not wall-clock replanning periods.

\noindent\textbf{Reference-attention statistics.}
AH uses attention from the conditional forward pass at denoising iteration
index 25 (zero-based) out of 50 updates. We select the 18 layers with
direct video-to-action connectivity, each with 32 heads; the six decoupled
layers do not contain direct reference keys and are excluded.
For action query $i$, layer $\ell$, and head $b$, let
$P^{\ell b}_{ir}$ denote the attention weight on visible key $r$.
Attention is computed after QK normalization, positional encoding, and
conditional-key concatenation, with softmax over \emph{all} visible keys
before reference weights are selected. Thus the reference competes with
other inputs for attention mass.
Let $\mathcal R$ contain all reference tokens and let $\mathcal R_g$
contain the reference tokens at latent time group $g$. The 13-frame video
encodes to $G=4$ latent time groups. Within each head, define
\begin{align}
  \mu_i^{\ell b}&=\sum_{r\in\mathcal R}P^{\ell b}_{ir},\\
  q_{ig}^{\ell b}&=\frac{\sum_{r\in\mathcal R_g}P^{\ell b}_{ir}}
                       {\mu_i^{\ell b}},\\
  E_i^{\ell b}&=-\frac{\sum_{g=1}^{G}q_{ig}^{\ell b}
                              \log q_{ig}^{\ell b}}{\log G}.
\end{align}
Spatial positions are summed before forming the reference-internal
temporal distribution $q$. Its entropy measures dispersion across four
latent time groups, not across the 13 raw frames. The statistics used for
selection are head/layer averages,
\begin{equation}
  (m_i,e_i)=\frac{1}{18\cdot32}
    \sum_{\ell,b}(\mu_i^{\ell b},E_i^{\ell b}).
\end{equation}
In particular, entropy is computed separately in each head before averaging.

\noindent\textbf{Execution-length selection.}
An action position is a truncation candidate when both $m_i<0.10$ and
$e_i>0.86$. We require three consecutive candidate positions \emph{within
the same prediction}.
If $j$ is the zero-based start of the first such interval, set
\begin{equation}
  h_k=\operatorname{clip}(j,6,24).
  \label{eq:real_ah_length}
\end{equation}
If no interval qualifies, set $h_k=24$. The rule seeks to execute only the
prefix before the candidate interval, subject to a minimum of six actions;
an interval starting before position 6 therefore still results in six
executed actions. AH is a deterministic heuristic without a separately
trained horizon predictor. Attention mass and entropy are not calibrated
task-success probabilities or confidence scores for reference following.

\noindent\textbf{Sampling and guidance.}
Both AH and no-AH inference use 50 Euler denoising updates. Action time
$t$ decreases linearly from 1 to 0, For the video stream, we apply a time shift of 5, giving
$t_v=5t/(1+4t)$; each stream is updated using its own adjacent time
increments. Each update evaluates a reference-conditioned branch and a
branch without reference video. Target-video flow uses reference guidance
\begin{equation}
  v^{v}_{\mathrm{guided}}=v^{v}_{\mathrm{no\text{-}ref}}
    +5\bigl(v^{v}_{\mathrm{ref}}-v^{v}_{\mathrm{no\text{-}ref}}\bigr).
\end{equation}
Action flow is taken directly from the reference-conditioned branch
(action CFG scale 1), without extrapolating the two action outputs.
The 50 updates thus use 100 denoising-model forward passes.
Evaluation uses an empty caption.

\noindent\textbf{AH ablation controls.}
The no-AH control fixes the execution horizon at $h_k=24$ instead of
selecting a 6--24-step prefix. The context advances by the executed duration.
Shorter prefixes increase observation acquisition and replanning per unit
of action time.

\subsection{Paired Dataset Construction and Split Checks}
\label{sec:paired_data_details}

\noindent\textbf{Episode-level split.}
A target robot episode can be paired with multiple human videos. We assign
all pairs that share a target episode to the same split and verify that
training and validation share no target episode identifiers.

\noindent\textbf{Synchronous human reenactment.}
Participants first practice following the source robot video, then record
while watching its playback and synchronizing the manipulation with it.
They use a viewpoint matched to the head camera and mimic the gripper's
grasping and release motions with their hands while avoiding dexterous finger
manipulations that the parallel gripper cannot reproduce.
All human demonstrations record only tabletop hand and forearm manipulations
without faces or personally identifiable information, and were collected with
participant consent.

\noindent\textbf{Keyframe editing and video generation.}
Source robot-video annotations, including Qwen3.8-generated descriptions,
provide the atomic action condition. Selected source frames are edited
with \texttt{gpt-image-2} using HaGRID
human-hand appearance references. The edit preserves the source scene,
objects, viewpoint, and contact relationships while replacing robot
manipulators with human hands and forearms.
Edited frames are quality-checked before serving as appearance references
for video editing with Seedance 2.5 or Wan 3.0.
The complete source robot clip provides the motion
reference, together with the action description and scene-preservation
instructions. No explicit robot-arm segmentation mask, depth map, or edge
map is supplied to the generator.
Accepted human clips retain the original robot observations and actions
as supervision.

\noindent\textbf{Cross-embodiment action supervision.}
Both Astribot and AgiBot recordings supply paired video--action training
examples. AgiBot examples retain their own recorded actions under the
shared action representation, with dimension-validity masks identifying
available action components. They contribute video and action supervision
during CT; real-robot deployment is evaluated on Astribot S1.

\noindent\textbf{VLM screening.}
We sample frames at the midpoints of equal-duration intervals within each
clip and present them to a vision--language model (VLM) in chronological
order. For paired comparisons, the source and generated frames occupy two
rows, with each column matching the same relative time in the two clips.

The visual-quality check tests for complete replacement of
visible robot manipulators, absence of residual robot geometry, anatomical
hand and forearm appearance, scene stability, and motion continuity.
Task tools already present in the source are distinguished from robot
residue. Skin-colored rigid robot links or gripper shapes fail the check
even when their surface appearance resembles skin.

For the source-alignment check, the verifier receives the paired frames
and the source action description. It checks action identity, left/right
roles, timing of the main phases, trajectories, object states, and absence
of additional manipulation. Different hand poses, incidental contact after
placement, and small differences in release or withdrawal timing are allowed
when the main manipulation and resulting object states agree.
Changing the target object, reversing a transport direction, or continuing
to manipulate an object after the source action ends fails this check.
The verifier returns a pass/fail decision for each check. We retain a
generated clip only if it passes both.

\end{document}